\documentclass[11pt]{article}
\usepackage{coling2020}
\usepackage{times}
\usepackage{url}
\usepackage{latexsym}
\usepackage{amssymb,booktabs,lipsum,xcolor,multirow,graphicx,subcaption,soul,wrapfig, blindtext}
\usepackage[ruled,linesnumbered]{algorithm2e}

\colingfinalcopy 

\title{An Analysis of Simple Data Augmentation for Named Entity Recognition}

\author{\begin{tabular}{cc}
Xiang Dai$^{1,2,3}$ & Heike Adel$^{1}$
\end{tabular}\\
\begin{tabular}{cc}
\multicolumn{2}{c}{$^{1}$Bosch Center for Artificial Intelligence, Renningen, Germany}\\
\multicolumn{2}{c}{$^{2}$University of Sydney, Sydney, Australia}\\
\multicolumn{2}{c}{$^{3}$CSIRO Data61, Sydney, Australia}\\
\tt dai.dai@csiro.au & \tt heike.adel@de.bosch.com \\
\end{tabular}
}
\date{}

\begin{document}
\maketitle
\begin{abstract}
Simple yet effective data augmentation techniques have been proposed for sentence-level and sentence-pair natural language processing tasks.
Inspired by these efforts, we design and compare data augmentation for named entity recognition, which is usually modeled as a token-level sequence labeling problem.
Through experiments on two data sets from the biomedical and materials science domains (i2b2-2010 and MaSciP), we show that simple augmentation can boost performance for both recurrent and transformer-based models, especially for small training sets.
\end{abstract}

\section{Introduction}
Modern deep learning techniques typically require a lot of labeled data~\cite{Bowman:Angeli:EMNLP:2015,Conneau:Schwenk:EACL:2017}.
However, in real-world applications, such large labeled data sets are not always available.
This is especially true in some specific domains, such as the biomedical and materials science domain, where annotating data requires expert knowledge and is usually time-consuming~\cite{Karimi:Metke:JBI:2015,Friedrich:Adel:ACL:2020}.
Different approaches have been investigated to solve this low-resource problem.
For example, transfer learning pretrains language representations on self-supervised or rich-resource source tasks and then adapts these representations to the target task~\cite{Ruder:Transfer:2019,Gururangan:Marasovic:ACL:2020}. 
Data augmentation expands the training set by applying transformations to training instances without changing their labels~\cite{Wang:Perez:arXiv:2017}.

Recently, there is an increased interest on applying data augmentation techniques on sentence-level and sentence-pair natural language processing (NLP) tasks, such as text classification~\cite{Wei:Zou:EMNLP:2019,Xie:Dai:arXiv:2019}, natural language inference~\cite{Min:McCoy:ACL:2020} and machine translation~\cite{Wang:Pham:EMNLP:2018}.
Augmentation methods explored for these tasks either create augmented instances by manipulating a few words in the original instance, such as word replacement~\cite{Zhang:Zhao:NIPS:2015,Wang:Yang:EMNLP:2015,Cai:Chen:ACL:2020}, random deletion~\cite{Wei:Zou:EMNLP:2019}, or word position swap~\cite{Sahin:Steedman:EMNLP:2018,Min:McCoy:ACL:2020}; or create entirely artificial instances via generative models, such as variational auto encoders~\cite{Yoo:Shin:AAAI:2019,Mesbah:Yang:EMNLP:2019} or back-translation models~\cite{Yu:Dohan:ICLR:2018,Iyyer:Wieting:NAACL:2018}.

Different from these sentence-level NLP tasks, named entity recognition (NER) does predictions on the token level. That is, for each token in the sentence, NER models predict a label indicating whether the token belongs to a mention and which entity type the mention has.
Therefore, applying transformations to tokens may also change their labels.
Due to this difficulty, data augmentation for NER is comparatively less studied.
In this work, we fill this research gap by exploring data augmentation techniques for NER, a token-level sequence labeling problem.


Our contributions can be summarized as follows:
\begin{enumerate}
    \item We survey previously used data augmentation techniques for sentence-level and sentence-pair NLP tasks and adapt some of them for the NER task.
    \item We conduct empirical comparisons of different data augmentations using two English domain-specific data sets: MaSciP~\cite{Mysore:Jensen:LAW:2019} and i2b2-2010~\cite{Uzuner:South:AMIA:2011}. Results show that simple augmentation can even improve over a strong baseline with large-scale pretrained transformers.
\end{enumerate}

\section{Related Work}
\label{section-related-work}
In this section, we survey previously used data augmentation methods for NLP tasks, grouping them into four categories:

\paragraph{Word replacement:}
Various word replacement variants have been explored for text classification tasks.
\newcite{Zhang:Zhao:NIPS:2015} and \newcite{Wei:Zou:EMNLP:2019} replace words with one of their synonyms, retrieved from an English thesaurus (e.g., WordNet).
\newcite{Kobayashi:NAACL:2018} replace words with other words that are predicted by a language model at the word positions.
\newcite{Xie:Dai:arXiv:2019} replace uninformative words with low TF-IDF scores with other uninformative words for topic classification tasks.

For machine translation, word replacement has also been used to generate additional parallel sentence pairs.
\newcite{Wang:Pham:EMNLP:2018} replace words in both the source and the target sentence by other words uniformly sampled from the source and the target vocabularies.
\newcite{Fadaee:Bisazza:ACL:2017} search for contexts where a common word can be replaced by a low-frequency word, relying on recurrent language models. 
\newcite{Gao:Zhu:ACL:2019} replace a randomly chosen word by a soft word, which is a probabilistic distribution over the vocabulary, provided by a language model.

%
In addition, there are two special word replacement cases, inspired by dropout and masked language modeling: replacing a word by a zero word (i.e., dropping entire word embeddings)~\cite{Iyyer:Manjunatha:ACL:2015}, or by a [MASK] token~\cite{Wu:Lv:arXiv:2018}.

\paragraph{Mention replacement:}
\newcite{Raiman:Miller:EMNLP:2017} augment a question answering training set using an external knowledge base. 
In particular, they extract nominal groups in the training set, perform string matching with entities in Wikidata, and then randomly replace them with other entities of the same type. In order to remove gender bias from coreference resolution systems, \newcite{Zhao:Wang:NAACL:2018} propose to generate an auxiliary dataset where all male entities are replaced by female entities, and vice versa, using a rule-based approach.

\paragraph{Swap words:}
\newcite{Wei:Zou:EMNLP:2019} randomly choose two words in the sentence and swap their positions to augment text classification training sets.
\newcite{Min:McCoy:ACL:2020} explore syntactic transformations (e.g., subject/object inversion) to augment the training data for natural language inference. \newcite{Sahin:Steedman:EMNLP:2018} rotate tree fragments around the root of the dependency tree to form a synthetic sentence and augment low-resource language part-of-speech tagging training sets. 

\paragraph{Generative models:}

\newcite{Yu:Dohan:ICLR:2018} train a question answering model with data generated by back-translation from a neural machine translation model.
\newcite{Kurata:Xiang:INTERSPEECH:2016} and \newcite{Hou:Liu:COLING:2018} use a sequence-to-sequence model to generate diversely augmented utterances to improve the dialogue language understanding module. \newcite{Xia:Kong:ACL:2019} convert data from a high-resource language to a low-resource language, using a bilingual dictionary and an unsupervised machine translation model in order to expand the machine translation training set for the low-resource language.




\section{Data Augmentation for NER~\label{section-augmentation-details}}

Inspired by the efforts described in Section~\ref{section-related-work}, we design several simple data augmentation methods for NER. Note that these augmentations do not rely on any externally trained models, such as machine translation models or syntactic parsing models, which are by themselves difficult to train in low-resource domain-specific scenarios.

\begin{table}[t]
	\centering
	\small
	\setlength{\tabcolsep}{1pt}
	\begin{tabular}{c|ccccc ccccc ccccc}
		\toprule
		& \multicolumn{12}{c}{\bf Instance} \\
		\midrule
		\multirow{2}{*}{None} & She & did & not & complain & of & headache & or & any & other & neurological & symptoms & . \\
		& O & O & O & O & O & B-problem & O & B-problem & I-problem & I-problem & I-problem & O \\
		\midrule
		\multirow{2}{*}{LwTR} & \textcolor{blue}{L.} & \textcolor{blue}{One} & not & complain & of & headache & \textcolor{blue}{he} & any & \textcolor{blue}{interatrial} & neurological & \textcolor{blue}{current} & . \\
		& O & O & O & O & O & B-problem & O & B-problem & I-problem & I-problem & I-problem & O \\
		\midrule
		\multirow{2}{*}{SR} & She & did & \textcolor{blue}{non} & complain & of & headache & or & \textcolor{blue}{whatsoever} & \textcolor{blue}{former} & neurologic & symptom & . \\
		& O & O & O & O & O & B-problem & O & B-problem & I-problem & I-problem & I-problem & O \\
		\midrule
		\multirow{2}{*}{MR} & She & did & not & complain & of & \textcolor{blue}{neuropathic} & \textcolor{blue}{pain} & \textcolor{blue}{syndrome} & or & \textcolor{blue}{acute} & \textcolor{blue}{pulmonary} & \textcolor{blue}{disease} & . \\
		& O & O & O & O & O & \textcolor{blue}{B-problem} & \textcolor{blue}{I-problem} & \textcolor{blue}{I-problem} & O & \textcolor{blue}{B-problem} & \textcolor{blue}{I-problem} & \textcolor{blue}{I-problem} & O \\
		\midrule
		\multirow{2}{*}{SiS} & \textcolor{blue}{not} & \textcolor{blue}{complain} & \textcolor{blue}{She} & \textcolor{blue}{did} & \textcolor{blue}{of} & headache & or & \textcolor{blue}{neurological} & \textcolor{blue}{any} & \textcolor{blue}{symptoms} & \textcolor{blue}{other} & . \\
		& O & O & O & O & O & B-problem & O & B-problem & I-problem & I-problem & I-problem & O \\
		\bottomrule 
	\end{tabular}
	\caption{Original training instance and different types of augmented instances. We highlight changes using \textcolor{blue}{blue} color. Note that LwTR (Label-wise token replacement) and SiS (Shuffle within segments) change token sequence only, whereas SR (Synonym replacement) and MR (Mention replacement) may also change the label sequence.}
	\label{table-augmentation-example}
\end{table}

\paragraph{Label-wise token replacement (LwTR):}
For each token, we use a binomial distribution to randomly decide whether it should be replaced. If yes, we then use a label-wise token distribution, built from the original training set, to randomly select another token with the same label. Thus, we keep the original label sequence unchanged. Taking the instance in Table~\ref{table-augmentation-example} as an example, there are five tokens replaced by other tokens which share the same label with the original tokens.

\paragraph{Synonym replacement (SR):}
Our second approach is similar to LwTR, except that we replace the token with one of its synonyms retrieved from WordNet. Note that the retrieved synonym may consist of more than one token. However, its BIO-labels can be derived using a simple rule: If the replaced token is the first token within a mention (i.e., the corresponding label is `B-EntityType'), we assign the same label to the first token of the retrieved multi-word synonym, and `I-EntityType' to the other tokens. If the replaced token is inside a mention (i.e., the corresponding label is `I-EntityType'), we assign its label to all tokens of the multi-word synonym.

\paragraph{Mention replacement (MR):}
For each mention in the instance, we use a binomial distribution to randomly decide whether it should be replaced. If yes, we randomly select another mention from the original training set which has the same entity type as the replacement.
The corresponding BIO-label sequence can be changed accordingly.
For example, in Table~\ref{table-augmentation-example}, the mention `headache [B-problem]' is replaced by another problem mention `neuropathic pain syndrome [B-problem I-problem I-problem]'.

\paragraph{Shuffle within segments (SiS):}
We first split the token sequence into segments of the same label. Thus, each segment corresponds to either a mention or a sequence of out-of-mention tokens. 
For example, the original sentence in Table~\ref{table-augmentation-example} is split into five segments: [She did not complain of], [headache], [or], [any other neurological symptoms], [.].
Then for each segment, we use a binomial distribution to randomly decide whether it should be shuffled. 
If yes, the order of the tokens within the segment is shuffled, while the label order is kept unchanged.

\paragraph{All:}
We also explore to augment the training set using all aforementioned augmentation methods. That is, for each training instance, we create multiple augmented instances, one per augmentation method.

\section{Experiments and Results}
\subsection{Datasets}
\begin{table}[b]
	\centering
	\begin{tabular}{r | ccc | ccc  }
		\toprule
		& \multicolumn{3}{c|}{\bf MaSciP} & \multicolumn{3}{c}{\bf i2b2-2010} \\
		\cline{2-7}
		& Train & Dev & Test & Train & Dev & Test \\
		\midrule
		Number of sentences & 1,901 & 109 & 158 & 13,868 & 2,447 & 27,625 \\ 
		Number of tokens & 61,750 & 4,158 & 4,585 & 129,087 & 20,454 & 267,249 \\ 
		Number of mentions & 18,874 & 1,190 & 1,259 & 14,376 & 2,143 & 31,161 \\ 
		Number of entity types & 21 & 20 & 21 & 3 & 3 & 3 \\
		\bottomrule
	\end{tabular}
	\caption{The descriptive statistics of the data sets.}
	\label{table-data-statistics}
\end{table}

We present an empirical analysis of the data augmentation methods described in Section~\ref{section-augmentation-details} on two English datasets from the materials science and biomedical domains: MaSciP~\cite{Mysore:Jensen:LAW:2019}\footnote{https://github.com/olivettigroup/annotated-materials-syntheses} and i2b2-2010~\cite{Uzuner:South:AMIA:2011}.\footnote{https://portal.dbmi.hms.harvard.edu/}

MaSciP contains synthesis procedures annotated with synthesis operations and their typed arguments (e.g., Material, Synthesis-Apparatus, etc.). 
We use the train-dev-test split provided by the authors.
i2b2-2010 focuses on the identification of Problem, Treatment and Test from patient reports.
We use the train-test split from its corresponding shared task setting and randomly select 15\% of sentences from the training set as the development set. 

To simulate a low-resource setting, we select the first 50, 150, 500 sentences which contain at least one mention from the training set to create the corresponding small, medium, and large training sets (denoted as S, M, L in Table~\ref{table-main-results}, whereas the complete training set is denoted as F) for each data set.
Note that we apply data augmentation only on the training set, without changing the development and test sets.

\subsection{Backbone models}
We model the NER task as a sequence-labeling task. 
Let $\mathbf{x}=\left\langle x_{1}, \ldots, x_{T}\right\rangle$ be a sequence of ${T}$ tokens, the model aims to predict a label sequence $\mathbf{y}=\left\langle y_{1}, \ldots, y_{T}\right\rangle$, where each label is composed of a position indicator
(e.g., BIO schema) and an entity type. 
The state-of-the-art sequence-labeling models roughly consist of two components: a neural-based encoder which creates contextualized embeddings $r_i$ for each token, and a conditional random field output layer, which captures dependencies between neighboring labels:
\[
\hat{P}(y_{1:T}|r_{1:T}) \propto \prod_{i=1}^T \psi_i (y_{i-1}, y_i, r_i).
\]

We consider two encoder variants in our study: one based on LSTM~\cite{Graves:Mohamed:ICASSP:2013} and one based on BERT~\cite{Devlin:Chang:NAACL:2019}.
The LSTM-based encoder consists of a context-independent token embedding layer (e.g., GloVe~\cite{Pennington:Socher:EMNLP:2014}) and a bidirectional LSTM layer, whose weights are learned from scratch.
The representations $r_i$ are obtained by concatenating the hidden states of the forward and backward LSTMs at each token position.
The BERT-based encoder consists of a sub-token embedding layer and a stack of multi-head self-attention and fully-connected feed-forward layers.
The final hidden state corresponding to the first sub-token within each token is used as the representation $r_i$.
Studies on domain-specific BERT models show that effectiveness on downstream tasks can be improved when the BERT models are further pretrained on in-domain data~\cite{Gururangan:Marasovic:ACL:2020,Dai:Karimi:EMNLP:2020}. We thus choose SciBERT~\cite{Beltagy:Lo:EMNLP:2019}, which is pretrained on scholar articles, and fine-tune it on the NER task.
In our preliminary experiments, we observe that SciBERT achieves significant better results than BERT~\cite{Devlin:Chang:NAACL:2019}.

We use the Micro-average string match $F_1$ score to evaluate the effectiveness of the models.
The model which is most effective on the development set, measured using the $F_1$ score, is finally evaluated on the test set. 

\subsection{Hyperparameters}
For each augmentation method, we tune the number of generated instances per training instance from a list of numbers: \{1, 3, 6, 10\}. When all data augmentation methods are applied, we reduce this tuning list to: \{1, 2, 3\}, so that the total number of generated instances given each original training instance is roughly the same for different experiments.
We also tune the $p$ value of the binomial distribution which is used to decide whether a token or a mention should be replaced (cf., Section~\ref{section-augmentation-details}). It is searched over the range from 0.1 to 0.7, with an incrementation step of 0.2. We perform grid search to find the best combination of these two hyperparameters on the developement set.

\begin{table}[t]
\begin{small}
    \setlength{\tabcolsep}{2pt}
    \centering
    \begin{tabular}{r | r | cccc | cccc | r}
    \toprule
    \multirow{2}{*}{\bf Model} & \multirow{2}{*}{\bf Method} & \multicolumn{4}{c|}{\bf MaSciP} & \multicolumn{4}{c|}{\bf i2b2-2010} & \multirow{2}{*}{\bf $\Delta$} \\
    \cline{3-10}
    & & S & M & L & F & S & M & L & F & \\ \midrule
    \multirow{7}{*}{Recurrent} & No augmentation & 53.0\scriptsize{$\pm$ 3.2} & 63.0\scriptsize{$\pm$ 0.6} & 70.3\scriptsize{$\pm$ 0.8} & 76.4\scriptsize{$\pm$ 0.4} & 17.1\scriptsize{$\pm$ 2.0} & 43.3\scriptsize{$\pm$ 1.2} & 54.1\scriptsize{$\pm$ 0.6} & 81.1\scriptsize{$\pm$ 0.2} & \\  
    & Label-wise token rep. & \underline{59.7\scriptsize{$\pm$ 0.6}} & \underline{65.5\scriptsize{$\pm$ 0.6}} & \underline{71.4\scriptsize{$\pm$ 0.4}} & 76.3\scriptsize{$\pm$ 0.8} & \underline{26.7\scriptsize{$\pm$ 0.8}} & \underline{44.3\scriptsize{$\pm$ 0.8}} & 54.5\scriptsize{$\pm$ 0.8} & 81.0\scriptsize{$\pm$ 0.2} & 2.6 \\  
    & Synonym replacement & \underline{60.1\scriptsize{$\pm$ 0.5}} & \underline{65.4\scriptsize{$\pm$ 0.4}} & 70.8\scriptsize{$\pm$ 0.6} & 76.7\scriptsize{$\pm$ 0.8} & \underline{25.9\scriptsize{$\pm$ 0.5}} & 44.1\scriptsize{$\pm$ 0.5} & 54.4\scriptsize{$\pm$ 1.5} & 81.0\scriptsize{$\pm$ 0.3} & 2.5 \\  
    & Mention replacement & \underline{60.6\scriptsize{$\pm$ 0.6}} & \underline{65.4\scriptsize{$\pm$ 0.4}} & \underline{71.9\scriptsize{$\pm$ 0.5}} & 76.0\scriptsize{$\pm$ 0.8} & \underline{25.9\scriptsize{$\pm$ 0.7}} & \underline{45.5\scriptsize{$\pm$ 0.4}} & \underline{55.0\scriptsize{$\pm$ 0.2}} & \underline{81.4\scriptsize{$\pm$ 0.2}} & 2.9 \\  
    & Shuffle within segments & \underline{58.8\scriptsize{$\pm$ 0.7}} & \underline{64.6\scriptsize{$\pm$ 0.4}} & 70.5\scriptsize{$\pm$ 0.8} & \underline{77.1\scriptsize{$\pm$ 0.3}} & \underline{25.2\scriptsize{$\pm$ 0.5}} & 44.4\scriptsize{$\pm$ 0.6} & 53.5\scriptsize{$\pm$ 0.9} & 80.6\scriptsize{$\pm$ 0.3} & 2.0 \\  
    & All & \underline{60.8\scriptsize{$\pm$ 1.3}} & \underline{67.0\scriptsize{$\pm$ 0.8}} & \underline{72.1\scriptsize{$\pm$ 0.7}} & 76.6\scriptsize{$\pm$ 0.4} & \underline{26.9\scriptsize{$\pm$ 0.7}} & \underline{45.4\scriptsize{$\pm$ 0.6}} & 54.6\scriptsize{$\pm$ 0.9} & \underline{81.5\scriptsize{$\pm$ 0.2}} & 3.3 \\  
    \midrule
    \multirow{7}{*}{Transformer} & No augmentation & 68.1\scriptsize{$\pm$ 0.6} & 72.7\scriptsize{$\pm$ 0.3} & 77.3\scriptsize{$\pm$ 0.5} & 79.8\scriptsize{$\pm$ 0.7} & 35.1\scriptsize{$\pm$ 1.1} & 62.7\scriptsize{$\pm$ 1.5} & 70.2\scriptsize{$\pm$ 0.3} & 87.8\scriptsize{$\pm$ 0.2} & \\ 
    & Label-wise token rep. & \underline{70.0\scriptsize{$\pm$ 0.8}} & 72.8\scriptsize{$\pm$ 0.2} & 76.0\scriptsize{$\pm$ 0.6} & 80.2\scriptsize{$\pm$ 0.6} & \underline{39.3\scriptsize{$\pm$ 1.7}} & 64.8\scriptsize{$\pm$ 1.3} & \underline{71.2\scriptsize{$\pm$ 0.4}} & 87.5\scriptsize{$\pm$ 0.2} & 1.0 \\  
    & Synonym replacement & \underline{70.6\scriptsize{$\pm$ 1.2}} & \underline{73.9\scriptsize{$\pm$ 0.1}} & 76.8\scriptsize{$\pm$ 0.4} & 79.7\scriptsize{$\pm$ 0.5} & \underline{42.3\scriptsize{$\pm$ 1.3}} & \underline{65.3\scriptsize{$\pm$ 0.3}} & 70.5\scriptsize{$\pm$ 2.3} & 87.7\scriptsize{$\pm$ 0.4} & 1.6 \\  
    & Mention replacement & \underline{70.5\scriptsize{$\pm$ 0.8}} & 73.3\scriptsize{$\pm$ 0.4} & 76.7\scriptsize{$\pm$ 0.7} & 80.0\scriptsize{$\pm$ 0.3} & \underline{40.1\scriptsize{$\pm$ 2.5}} & 64.2\scriptsize{$\pm$ 1.2} & 70.8\scriptsize{$\pm$ 0.7} & 87.8\scriptsize{$\pm$ 0.2} & 1.2 \\  
    & Shuffle within segments & \underline{70.5\scriptsize{$\pm$ 0.4}} & 73.1\scriptsize{$\pm$ 0.6} & 76.7\scriptsize{$\pm$ 0.3} & 80.3\scriptsize{$\pm$ 0.5} & \underline{39.4\scriptsize{$\pm$ 1.6}} & 63.9\scriptsize{$\pm$ 1.4} & 71.2\scriptsize{$\pm$ 1.2} & 87.7\scriptsize{$\pm$ 0.2} & 1.1 \\  
    & All & \underline{71.2\scriptsize{$\pm$ 0.8}} & 73.1\scriptsize{$\pm$ 0.6} & 76.9\scriptsize{$\pm$ 0.4} & 80.5\scriptsize{$\pm$ 0.4} & \underline{41.5\scriptsize{$\pm$ 0.9}} & \underline{65.2\scriptsize{$\pm$ 0.3}} & \underline{72.3\scriptsize{$\pm$ 1.3}} & 87.2\scriptsize{$\pm$ 0.3} & 1.8 \\  
    \bottomrule
    \end{tabular}
    \caption{Evaluation results in terms of span-level F1 score. \textbf{S}mall set contains 50 training instances; \textbf{M}edium contains 150 instances; \textbf{L}arge contains 500 instances; \textbf{F}ull uses the complete training set. We repeat all experiments five times with different random seeds. Mean values and standard deviations are reported. $\Delta$ column shows the averaged improvement due to data augmentation. \underline{underline}: the result is significantly better than the baseline model without data augmentation (paired student's t-test, p: $0.05$). }
    \label{table-main-results}
\end{small}
\end{table}

\subsection{Results}
Table~\ref{table-main-results} provides the evaluation results on the test sets.
The first conclusion we can draw is that all data augmentation techniques can improve over the baseline where no augmentation is used, although there is no single clear winner across both recurrent and transformer models. Synonym replacement outperforms other augmentation on average when transformer models are used, whereas mention replacement appears to be most effective for recurrent models.

Second, applying all data augmentation methods together outperforms any single data augmentation on average, although, when the complete training set is used, applying single data augmentation may achieve better results (c.f., MaSciP-Recurrent and i2b2-2010-Transformer). This scenario may reflect a trade-off between diversity and validity of augmented instances~\cite{Hou:Liu:COLING:2018,Xie:Dai:arXiv:2019}. On the one hand, applying all data augmentation together may prevent overfitting via producing diverse training instances. This positive effect is especially useful when the training sets are small. On the other hand, it may also increase the risk of altering the ground-truth label, or generating invalid instances. This negative effect may dominate for larger training sets.

Third, data augmentation techniques are more effective when the training sets are small. For example, all data augmentation methods achieve significant improvements when the training set contains only 50 instances. In contrast, when the complete training sets are used, only three augmentation methods achieve significant improvements and some even decrease the performance. This has also been observed in previous work on machine translation tasks~\cite{Fadaee:Bisazza:ACL:2017}.

Last but not least, we notice that previous studies mainly investigate the effectiveness of data augmentation with recurrent models where most of the parameters are learned from scratch. Considering the significant improvements when using pretrained transformer models, we argue that it is important to investigate the effectiveness of techniques also on pretrained models, such as BERT~\cite{Devlin:Chang:NAACL:2019}, because they are supposed to capture various knowledge via self-supervision learning.

\section{Conclusion}

We survey previously used data augmentation methods for sentence-level and sentence-pair NLP tasks and adapt them to NER, a token-level task. Through experiments on two domain-specific data sets, we show that simple data augmentation can improve performance even over strong baselines. 
\bibliographystyle{coling}
\bibliography{coling2020}


\end{document}